# Stiffness Analysis of Overconstrained Parallel Manipulators

Anatol Pashkevich[1,2], Damien Chablat[1*], Philippe Wenger[1]

[1]*Institut de Recherches en Communications et Cybernétique de Nantes,*
*UMR CNRS 6597, 1 rue de la No, 44321 Nantes, France*

[2]*Ecole des Mines de Nantes,*
*4 rue Alfred-Kastler, Nantes 44307, France*

**Abstract**. The paper presents a new stiffness modeling method for overconstrained parallel manipulators with flexible links and compliant actuating joints. It is based on a multidimensional lumped-parameter model that replaces the link flexibility by localized 6-dof virtual springs that describe both translational/rotational compliance and the coupling between them. In contrast to other works, the method involves a FEA-based link stiffness evaluation and employs a new solution strategy of the kinetostatic equations for the unloaded manipulator configuration, which allows computing the stiffness matrix for the overconstrained architectures, including singular manipulator postures. The advantages of the developed technique are confirmed by application examples, which deal with comparative stiffness analysis of two translational parallel manipulators of 3-PUU and 3-PRPaR architectures. Accuracy of the proposed approach was evaluated for a case study, which focuses on stiffness analysis of Orthoglide parallel manipulator.

## 1. Introduction

Parallel manipulators have become more and more popular in industrial applications, including high-accuracy positioning and high-speed machining [1, 2]. This growing attention is inspired by their essential advantages over serial manipulators, which have already reached the dynamic performance limits (bounded by high masses of the machine components required to support sequential joints, links and actuators). In contrast, parallel manipulators are claimed to offer better accuracy, lower mass/inertia properties, and higher structural rigidity (i.e. stiffness-to-mass ratio) [3]. These features are induced by their specific kinematic structure, which resists the error accumulation in kinematic chains and allows convenient actuators location close to the manipulator base. Besides, the links act in parallel against the external force/torque, eliminating the cantilever-type loading and increasing the manipulator stiffness [4]. The latter makes them attractive for innovative machine-tool architectures [5-7], but practical utilization of the potential benefits requires development of efficient stiffness analysis techniques, which satisfy the computational speed and accuracy requirements of relevant design procedures [8].

Generally, the stiffness analysis evaluates the effect of the applied external torques and forces on the compliant displacements of the end-effector. Numerically, this property is defined through the "stiffness matrix" $K$, which gives the relation between the translational/rotational displacement and the static forces/torques causing this transition. The inverse of $K$ is usually called the "compliance matrix" and is denoted as $k$. As follows from mechanics, $K$ is 6×6 semi-definite non-negative matrix, where structure may be non-diagonal to represent the coupling between the translation and rotation [9]. Besides, this matrix may be not-symmetrical under the static load [10], but standard stiffness analysis focuses on the non-loaded structures.

---

\* Corresponding author: Tel: +33 240 37 69 48; Fax: +33 240 37 69 30;
E-mail: Damien.Chablat@irccyn.ec-nantes.fr

Similar to other manipulator properties (kinematical, for instance), the stiffness essentially depends on the force/torque direction and on the manipulator configuration. Hence, to provide the designer with integrated performance criteria, various scalar indices are usually computed (such as the best/worst/average stiffness with respect to the rotation or translation). They are typically derived using the singular-value decomposition of $K$. However, there are still a number of open questions here regarding the significance of these indices for a particular manufacturing task. Besides, since the matrix $K$ varies through the workspace, corresponding global benchmarks must be computed. In some cases, a relevant analysis produces the "stiffness maps", which describe the end-effector compliance as a function of the manipulator configuration [11-13].

Several approaches exist for the computation of the stiffness matrix, which differ in the modeling assumptions and computational techniques. They are the Finite Element Analysis (FEA), the matrix structural analysis (MSA), and the virtual joint method (VJM) that is often called the lumped modeling.

The FEA method is proved to be the most accurate and reliable, since the links/joints are modeled with its true dimension and shape [14]. Its accuracy is limited by the discretisation step only. However, because of high computational expenses required for the repeated re-meshing, this method is usually applied at the final design stage for the verification and component dimensioning. For example, in [15], a FEA model was used to evaluate the static rigidity and natural frequencies of the T3R1 parallel robot. Also, this method is widely used for validation of other stiffness analysis techniques [16-18] and for the comparative study [19].

The MSA method is a common technique in mechanical engineering [20], it incorporates the main ideas of the FEA but operates with rather large flexible elements (beams, arcs, cables, etc.). This obviously yields reduction of the computational expenses and, in some cases, allows even obtaining an analytical stiffness matrix. For parallel manipulators, the relevant stiffness model is a combination of flexible beams and nodes, where each beam is defined by two nodes and described by 12×12 stiffness matrix derived from the Euler-Bernoulli presentation. Then, these matrices are "assembled" in accordance with the superposition principle and the manipulator geometry, to produce the desired 6×6 matrix for the whole mechanism. Sometimes this approach is also referred to as the "distributed stiffness" modeling. One of the first examples of MSA application for the problem of interest is the stiffness analysis of a Stewart platform [21], which was performed under the assumption that the links are not subject to bending. This approach was also used in [22, 23] for other manipulators and/or other modeling assumptions. Some resent MSA-based results are obtained for the Delta-type mechanisms [24]. This method gives a reasonable trade-off between the accuracy and computational time, provided that link approximation by the beam elements is realistic. Because it involves rather high-dimensional matrix operations, it is not attractive for the parametric stiffness analysis and analytical modeling.

Finally, the VJM method, which is also referred to as the "lumped modeling", is based on the expansion of the traditional rigid model by adding virtual joints (localized springs), which describe the elastic deformations of the manipulator components (links, joints and actuators).

This approach originates from the work of Gosselin [25], who evaluated parallel manipulator stiffness taking into account only the actuators compliance and by presenting them as one-dimensional linear springs (the links were assumed to be rigid, and the passive joints to be perfect). Besides, the compliance in all actuated joints was assumed to be equal. The latter allowed reducing the stiffness analysis to the analysis of the condition number of the Jacobian matrix. Further development of VJM allowed taking into account the links flexibility, which were presented as rigid beams supplemented by linear and torsional springs [26]. There are a number of variations and simplifications of the VJM method, which differ in modeling assumptions and numerical techniques. In particular, it was applied to the CaPAMan, Orthoglide and H4 robots, specific variants of Stewart-Gough platform, manipulators with US/UPS legs and other kinematic machines [27-32]. Generally, the lumped modeling provides acceptable accuracy in short computational time, so it is widely used at the pre-design stage, especially for the analytical parametric analysis. However, it is very hypothetic and operates with simplified stiffness models that are composed of one-dimensional springs that do not take into account the coupling between the rotational and translational deflections. There are also other restrictions, which limit its applications to non-overconstrained mechanisms.

This paper presents a new stiffness modelling method, which combines advantages of the above mentioned approaches. It is based on a multidimensional lumped-parameter model that replaces the link flexibility by localized 6-dof virtual springs that describe both the linear/rotational deflections and the coupling between them. The spring stiffness parameters are evaluated using FEA modelling to ensure higher accuracy. In addition, it employs a new solution strategy of the kinetostatic equations, which allows computing the stiffness matrix for the overconstrained architectures, including the singular manipulator postures. This gives almost the same accuracy as FEA but with essentially lower computational effort because it eliminates the model re-meshing through the workspace.

Since the developed technique is targeted to the design optimization, it relies on the assumption that the manipulator is located in unloaded equilibrium configuration. This allows evaluating the symmetrical part of the general stiffness matrix while neglecting the skew-symmetrical components, which describe effects caused by the external loading and relevant changes in Jacobian [12, 33, 34]. It is obvious that at the preliminary design stage, which focuses on the conceptual issues (such as comparison of alternative manipulator architectures, defining critical components in the kinematic chains, deciding on the stiffness specifications for the links, etc.), this assumption is practical and reasonable. Besides, for a particular case study presented below, validity of the assumption is justified numerically.

The remainder of this paper is organized as follows. Next section introduces a general methodology of deriving/computing of the kinematic and stiffness model. Section 3 describes the manipulator compliant elements and proposes FEA-based technique for the evaluating their parameters. Section 4 includes application examples, which deal with comparative stiffness analysis of two translational parallel manipulators and demonstrate advantages of the proposed technique. Section 5 summarizes the main contributions of this work and defines future research directions.

## 2. General methodology

### 2.1. *Manipulator Architecture*

Let us consider a general *n*-dof parallel manipulator, which consists of a mobile platform connected to a fixed base by *n* identical kinematics chains (Fig. 1). Each chain includes an actuated joint "Ac" (prismatic or rotational) followed by a "Foot" and a "Leg" with a number of passive joints "Ps" inside. Generally, certain geometrical conditions are assumed to be satisfied with respect to the passive joints to eliminate the undesired platform rotations and to achieve stability of desired motions.

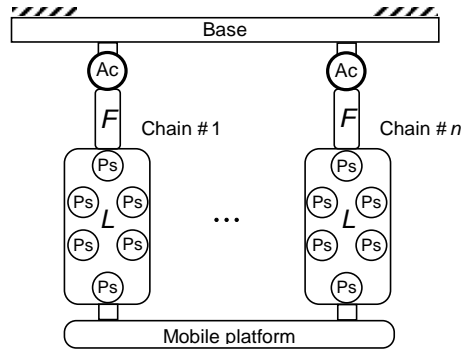

Fig. 1. Schematic diagram of a general n-dof parallel manipulator
(Ac – actuated joint, Ps – passive joints, F – foot, L - Leg)

Typical examples of such architectures are:

(a) 3-PUU translational parallel kinematic machine (Fig 2a); where each leg consists of a rod ended by two U-joints (with parallel intermediate and exterior axes), and active joint is driven by linear actuator [35];

(b) Delta parallel robot (Fig 2b) that is based on the 3-RRPaR architecture with parallelogram-type legs and rotational active joints [36];

(c) Orthoglide parallel robot (Fig 2c) that implements the 3-PRPaR architecture with parallelogram-type legs and translational active joints [37].

Here R, P, U and Pa denote the revolute, prismatic, universal and parallelogram joints, respectively.

It should be noted that examples (b) and (c) illustrate specific cases of the overconstrained mechanisms, for which the standard stiffness analysis methods cannot be applied directly. In particular, for the Orthoglide mechanism the architectural particularities can be summarized as follows: (i) each kinematic chain prevents the platform from rotating about two orthogonal axes; (ii) any combination of two kinematic chains suppresses the three platform rotations, (iii) the whole set of three kinematic chains also suppresses the three platform rotations. Hence, the kinematical architecture of this manipulator includes excessive (redundant) constrains that are in certain agreement for the nominal parameter values. However, such a spatial overconstrained[1] arrangement ensures essential increase of the rigidity with respect to the external force. This motivates development of dedicated stiffness analysis techniques that are presented below.

---

[1] In the robotic literature, an alternative definition of the "overconstrained manipulator" is also used. Some authors [39] use this the term in wide-sense, as the opposite to the "*redundant manipulator*", in order to distinguish mechanisms with any additional kinematic chains or specific location/orientation of the joints that cause reduction of the workspace dimension (down to planar, spherical, etc.). In this paper, the term "overconstrained" is referred to as the specific case, when the number of imposed constrains is greater that the resulting loss in the degrees-of-freedom [40]; it is also known as the "repeated" or "common"

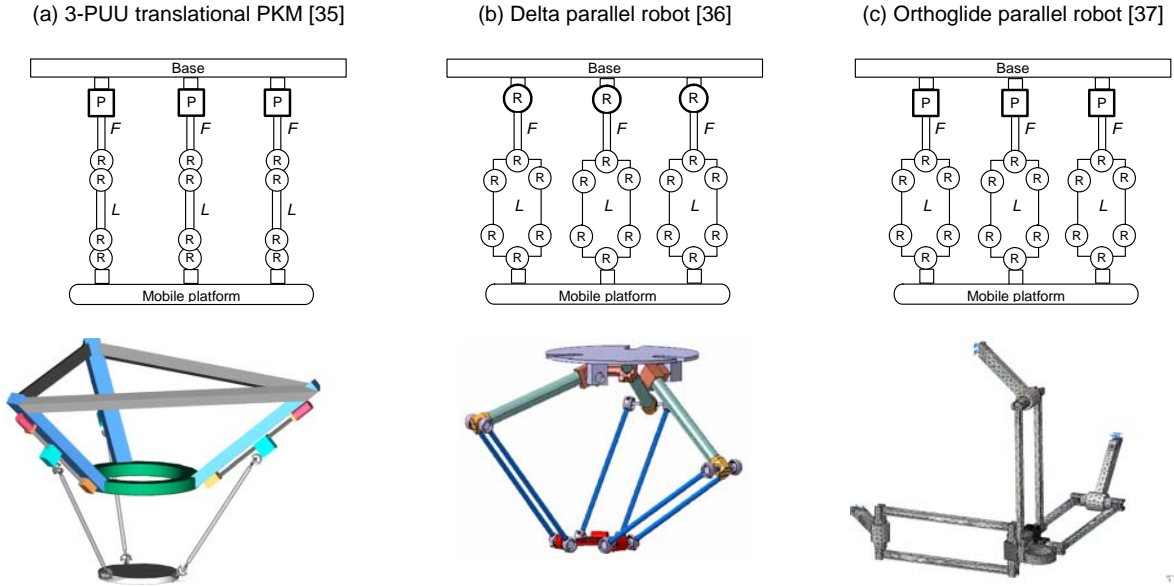

Fig. 2. Typical 3 dof translational parallel mechanisms

## 2.2. Basic Assumptions

To evaluate the manipulator stiffness, let us apply a modification of the virtual joint method (VJM), which is based on the lump modeling approach [25, 26]. According to this approach, the original rigid model should be extended by adding virtual joints (localized springs), which describe elastic deformations of the links. Besides, virtual springs are included in the actuating joints to take into account stiffness of the mechanical transmissions and the control loop. To overcome difficulties with parallelogram stiffness modeling, let us first replace the manipulator legs (see Fig. 2) by rigid links with configuration-dependent stiffness. (Such transformation will be justified further while deriving the stiffness model for the parallelogram-based legs).

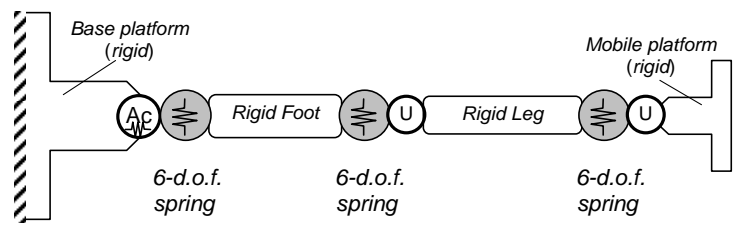

Fig. 3. Flexible model of a single kinematic chain (Ac – actuating joint, U – universal joint)

This transforms the general architecture into the extended *n-x*UU case allowing treating all the considered manipulators in a similar manner. Under such assumptions, each kinematic chain of the manipulator can be described by a serial structure (Fig. 3), which includes sequentially:

(a) a rigid link between the manipulator *base* and the *i*th actuating joint (part of the base platform) described by the constant homogenous transformation matrix $\mathbf{T}_{Base}^{i}$;

(b) a 1-dof actuating joint with supplementary virtual spring describing the *control loop stiffness*, which is defined by the homogenous matrix function $\mathbf{V}_a(q_0^i + \theta_0^i)$ where $q_0^i$ is the actuated coordinate and $\theta_0^i$ is the virtual spring coordinate;

(c) a 6-dof virtual spring describing the *actuator mechanical stiffness*, which is defined by the homogenous matrix function $\mathbf{V}_s(\theta_1^i,...\theta_6^i)$ where $\{\theta_1^i, \theta_2^i, \theta_3^i\}$, $\{\theta_4^i, \theta_5^i, \theta_6^i\}$ are the virtual spring coordinates corresponding to the spring translational and rotational deflections;

(d) a rigid "Foot" linking the actuating joint and the leg, which is described by the constant homogenous transformation matrix $\mathbf{T}_{Foot}$;

(e) a 6-dof virtual spring describing the *foot stiffness*, which are defined by the homogenous matrix function $\mathbf{V}_s(\theta_7^i,...\theta_{12}^i)$, where $\{\theta_7^i, \theta_8^i, \theta_9^i\}$ and $\{\theta_{10}^i, \theta_{11}^i, \theta_{12}^i\}$ are the spring translational/rotational deflections;

(f) a 2-dof passive U-joint at the beginning of the leg allowing two independent rotations with angles $\{q_1^i, q_2^i\}$, which is described by the homogenous matrix function $\mathbf{V}_{u1}(q_1^i, q_2^i)$;

(g) a rigid "Leg" linking the foot to the movable platform, which is described by the constant homogenous matrix transformation $\mathbf{T}_{Leg}$;

(h) a 6-dof virtual spring describing the *leg stiffness*, which are defined by the homogenous matrix function $\mathbf{V}_s(\theta_{13}^i,...\theta_{18}^i)$, where $\{\theta_{13}^i, \theta_{14}^i, \theta_{15}^i\}$ and $\{\theta_{16}^i, \theta_{17}^i, \theta_{18}^i\}$ correspond to the spring translations and rotations respectively;

(i) a 2-dof passive U-joint at the end of the leg allowing two independent rotations with angles $\{q_3^i, q_4^i\}$, which is described by the homogenous matrix function $\mathbf{V}_{u2}(q_3^i, q_4^i)$;

(j) a rigid link from the manipulator leg the *end-effector* (part of the movable platform) described by the constant homogenous matrix transformation $\mathbf{T}_{Tool}^i$.

The corresponding mathematical expression defining the end-effector location subject to variations of all above defined coordinates of a single kinematic chain *i* may be written as follows

$$\mathbf{T}_i = \mathbf{T}_{Base}^i \cdot \mathbf{V}_a(q_0^i + \theta_0^i) \cdot \mathbf{V}_s(\theta_1^i,...\theta_6^i) \cdot \mathbf{T}_{Foot} \cdot \mathbf{V}_s(\theta_7^i,...\theta_{12}^i) \cdot \mathbf{V}_{u1}(q_1^i, q_2^i) \cdot \mathbf{T}_{Leg} \cdot \mathbf{V}_s(\theta_{13}^i,...\theta_{18}^i) \cdot \mathbf{V}_{u2}(q_3^i, q_4^i) \cdot \mathbf{T}_{Tool}^i \quad (1)$$

where the matrix function $\mathbf{V}_a(.)$ is either an elementary rotation or translation, the matrix functions $\mathbf{V}_{u1}(.)$ and $\mathbf{V}_{u2}(.)$ are compositions of two successive rotations, the spring matrix $\mathbf{V}_s(.)$ is composed of six elementary transformations, and $i=1,...n$. In the rigid case, the virtual joint coordinates $\theta_1^i,...\theta_{18}^i$ are equal to zero, while the remaining ones (both active $q_0^i$ and passive $q_1^i,...q_4^i$) are obtained through the inverse kinematics, ensuring that all *n* matrices $\mathbf{T}_i$, $i=1,...n$ are equal to the prescribed one, that characterizes the desired spatial location of the moving platform (kinematic loop-closure equations). Particular expressions for all components of the product (1) may be easily derived using standard techniques for homogenous transformation matrices.

It should be noted that the kinematic model (1) includes 24 variables (1 for active joint, 4 for passive joints, and 19 for virtual springs). However, some of the virtual springs are redundant, since they are compensated by corresponding passive joints (with aligning axes) or by combination of passive joints. For computational convenience, nevertheless, it is not reasonable to detect and analytically eliminate redundant variables at this step, because the technique developed below allows easy and efficient computational elimination (without increasing the size of the required matrix inverse, which is equal to 6×6 independent of the virtual joint number).

## 2.3. Differential kinematic model

To evaluate the manipulator ability to respond to external forces and torques, let us first derive the differential kinematic equation describing relations between the end-effector location and small variations in the joint variables. For each $i$th kinematic chain, this equation can be generalized as follows

$$\delta \mathbf{t}_i = \mathbf{J}_\theta^i \cdot \delta \boldsymbol{\theta}_i + \mathbf{J}_q^i \cdot \delta \mathbf{q}_i, \quad i=1,...n, \tag{2}$$

where vector $\delta \mathbf{t}_i = (\delta p_{xi}, \delta p_{yi}, \delta p_{zi}, \delta \varphi_{xi}, \delta \varphi_{yi}, \delta \varphi_{zi})^T$ describes the end-effector translation $\delta \mathbf{p}_i = (\delta p_{xi}, \delta p_{yi}, \delta p_{zi})^T$ and rotation $\delta \boldsymbol{\varphi}_i = (\delta \varphi_{xi}, \delta \varphi_{yi}, \delta \varphi_{zi})^T$ with respect to the Cartesian axes; vector $\delta \boldsymbol{\theta}_i = (\delta \theta_0^i, ... \delta \theta_{18}^i)^T$ collects all virtual joint coordinates, the vector $\delta \mathbf{q}_i = (\delta q_1^i, ... \delta q_4^i)^T$ includes all passive joint coordinates, symbol '$\delta$' stands for variation with respect to the rigid case values, and $\mathbf{J}_\theta^i$, $\mathbf{J}_q^i$ are matrices of sizes 6x19 and 6x4 correspondingly. It should be noted that the derivative for the actuated coordinate $q_0^i$ is not included in $\mathbf{J}_q^i$ but it is represented in the first column of $\mathbf{J}_\theta^i$ through the variable $\theta_0^i$.

The desired matrices $\mathbf{J}_\theta^i$, $\mathbf{J}_q^i$, which are the only parameters of the differential model (2) may be computed from (1) analytically using some software support tools such as Maple, MathCAD or Mathematica. However, a straightforward differentiation usually yields very awkward expressions that are not convenient for further computations. On the other hand, the fractionized structure of (1), where all variables are separated, allows applying an efficient semi-analytical method. To present this technique, let us assume that for the particular virtual joint variable $\theta_j^i$ the model (1) is rewritten as

$$\mathbf{T}_i = \mathbf{H}_{ij}^L \cdot \mathbf{V}_{\theta j}(\theta_j^i) \cdot \mathbf{H}_{ij}^R, \tag{3}$$

where the first and the third multipliers are the constant homogenous matrices, and the second multiplier is the elementary translation or rotation. Then the partial derivative of the homogenous matrix $\mathbf{T}_i$ with respect to $\theta_j^i$ at the point $\theta_j^i = 0$ may be computed from a similar product where the internal term is replaced by the matrix $\mathbf{V}'_{\theta j}(.)$ that admits a very simple analytical presentation. In particular, for the elementary translations and rotations about the X-axis these derivatives are:

$$\mathbf{V}'_{Tran_x} = \begin{bmatrix} 0 & 0 & 0 & 1 \\ 0 & 0 & 0 & 0 \\ 0 & 0 & 0 & 0 \\ 0 & 0 & 0 & 0 \end{bmatrix}; \quad \mathbf{V}'_{Rot_x} = \begin{bmatrix} 0 & 0 & 0 & 0 \\ 0 & 0 & 1 & 0 \\ 0 & -1 & 0 & 0 \\ 0 & 0 & 0 & 0 \end{bmatrix}. \tag{4}$$

Furthermore, since the derivative of the homogenous matrix $\mathbf{T}'_i = \mathbf{H}_{ij}^L \cdot \mathbf{V}'_{\theta j}(\theta_j^i) \cdot \mathbf{H}_{ij}^R$ may be presented as

$$\mathbf{T}'_i = \left[ \begin{array}{ccc|c} 0 & \varphi'_{iz} & -\varphi'_{iy} & p'_{ix} \\ -\varphi'_{iz} & 0 & \varphi'_{ix} & p'_{iy} \\ \varphi'_{iy} & -\varphi'_{ix} & 0 & p'_{iz} \\ \hline 0 & 0 & 0 & 0 \end{array} \right], \tag{5}$$

So, the desired *j*th column of the jacobian $\mathbf{J}_\theta^i$ can be easily extracted from $\mathbf{T}_i'$ (using the matrix elements $T_{14}'$, $T_{24}'$, $T_{34}'$, $T_{23}'$, $T_{31}'$, $T_{12}'$).

The jacobians $\mathbf{J}_q^i$ can be computed in a similar manner, but the derivatives are evaluated in the neighborhood of the "nominal" values of the passive joint coordinates $q_{j\,nom}^i$ corresponding to the rigid case (these values are obtained from the inverse kinematics). However, a simple transformation $q_j^i = q_{j\,nom}^i + \delta q_j^i$ and a corresponding factoring of the function $\mathbf{V}_{qj}(q_j^i) = \mathbf{V}_{qj}(q_{j\,nom}^i) \cdot \mathbf{V}_{qj}(\delta q_j^i)$ allow applying the above approach. It is also worth mentioning that this technique may be also applied in analytical computations, allowing one to avoid bulky transformations required for the straightforward differentiating.

### 2.4. *Kinetostatic and Stiffness Models*

For the manipulator kinetostatic model that describes the force-and-motion relation, it is necessary to introduce additional equations that define the virtual joint reactions to the corresponding spring deformations. In accordance with the adopted stiffness model, the following virtual springs are included in each kinematic chain:

- 1-dof virtual spring describing the actuator control loop compliance;
- 6-dof virtual spring describing the actuator mechanics compliance;
- 6-dof virtual spring describing mechanical compliance of the foot;
- 6-dof virtual spring describing mechanical compliance of the leg.

Assuming that the spring deformations are small enough, the required relations may be expressed by linear equations

$$\left[\tau_{\theta 0}^i\right] = K_{ctr}\left[\theta_0^i\right]; \quad \begin{bmatrix} \tau_{\theta 1}^i \\ \vdots \\ \tau_{\theta 6}^i \end{bmatrix} = \mathbf{K}_{act}\begin{bmatrix} \theta_1^i \\ \vdots \\ \theta_6^i \end{bmatrix}; \quad \begin{bmatrix} \tau_{\theta 7}^i \\ \vdots \\ \tau_{\theta 12}^i \end{bmatrix} = \mathbf{K}_{Foot}\begin{bmatrix} \theta_7^i \\ \vdots \\ \theta_{12}^i \end{bmatrix}; \quad \begin{bmatrix} \tau_{\theta 13}^i \\ \vdots \\ \tau_{\theta 18}^i \end{bmatrix} = \mathbf{K}_{Leg}\begin{bmatrix} \theta_{13}^i \\ \vdots \\ \theta_{18}^i \end{bmatrix}, \quad (6)$$

where $\tau_{\theta j}^i$ is the generalized force for the *j*th virtual joint of the *i*th kinematic chain, $K_{ctr}$ is the actuator control loop stiffness (scalar), and $\mathbf{K}_{act}$, $\mathbf{K}_{Foot}$, $\mathbf{K}_{Leg}$ are 6×6 stiffness matrices for the mechanics of the actuator, foot and leg respectively. It should be stressed that, in contrast to other works, these matrices are assumed to be non-diagonal. This allows taking into account complicated coupling between rotational and translational deformations, while usual lump-based approach does not consider this phenomena [25]. Some examples of such compliance matrices will be given in subsequent sections.

For analytical convenience, expressions (6) may be collected in a single matrix equation

$$\boldsymbol{\tau}_\theta^i = \mathbf{K}_\theta \cdot \delta\boldsymbol{\theta}_i, \quad i = 1,...n \quad (7)$$

where $\boldsymbol{\tau}_\theta^i = (\tau_{\theta 0}^i, \ldots \tau_{\theta 18}^i)^T$ is the aggregated vector of the virtual joint reactions, and $\mathbf{K}_\theta = diag(K_{ctr}, \mathbf{K}_{act}, \mathbf{K}_{Foot}, \mathbf{K}_{Leg})$ is the aggregated spring stiffness matrix of size 19×19. Similarly, one can define the aggregated vector of the passive joint reactions $\boldsymbol{\tau}_q^i = (\tau_{q1}^i, \ldots \tau_{q4}^i)^T$ but all its components must be equal to zero:

$$\boldsymbol{\tau}_q^i = \mathbf{0}, \quad i = 1,...n. \tag{8}$$

To find the static equations corresponding to the end-effector motion $\delta \mathbf{t}_i$, let us apply the principle of virtual work assuming that the joints are given small, arbitrary virtual displacements $(\Delta \boldsymbol{\theta}_i, \Delta \mathbf{q}_i)$ in the equilibrium neighborhood. Then the virtual work of the external force $\mathbf{f}_i$ applied to the end-effector along the corresponding displacement $\Delta \mathbf{t}_i = \mathbf{J}_\theta^i \cdot \Delta \boldsymbol{\theta}_i + \mathbf{J}_q^i \cdot \Delta \mathbf{q}_i$ is equal to the sum $(\mathbf{f}_i^T \mathbf{J}_\theta^i) \cdot \Delta \boldsymbol{\theta}_i + (\mathbf{f}_i^T \mathbf{J}_q^i) \cdot \Delta \mathbf{q}_i$. For the internal forces, the virtual work is $-\boldsymbol{\tau}_\theta^{iT} \cdot \Delta \boldsymbol{\theta}_i$ since the passive joints do not produce the force/torque reactions (the minus sign takes into account the adopted directions for the virtual spring forces/torques). Therefore, because in the static equilibrium the total virtual work is equal to zero for any virtual displacement, the equilibrium conditions may be written as

$$\mathbf{J}_\theta^{iT} \cdot \mathbf{f}_i = \boldsymbol{\tau}_\theta^i; \qquad \mathbf{J}_q^{iT} \cdot \mathbf{f}_i = \mathbf{0}. \tag{9}$$

This gives additional expressions describing the force/torque propagation from the joints to the end-effector.

Hence, the complete kinetostatic model consists of five matrix equations (2), (7)…(9) where either $\mathbf{f}_i$ or $\delta \mathbf{t}_i$ are treated as known, and the remaining variables are considered as unknowns. Obviously, since separate kinematic chains posses some degrees-of-freedom, this system cannot be uniquely solved for given $\mathbf{f}_i$. However, vice versa, for given end-effector displacement $\delta \mathbf{t}_i$, it is possible to compute both the corresponding external force $\mathbf{f}_i$ and the internal variables, $\delta \boldsymbol{\theta}_i$, $\boldsymbol{\tau}_\theta^i$, $\delta \mathbf{q}_i$ (i.e. virtual spring reactions and displacements in passive joints, which may also provide useful information for the designer).

Using the above equations, the desired Cartesian stiffness matrix may be derived in a straightforward way, by differentiating (9) and relevant eliminating the redundant variables. In general case, this produces the stiffness matrix that consists of two components: (i) the symmetrical part, which describes the manipulator intrinsic stiffness properties in the neighborhood of the "unloaded equilibrium" (i.e. reaction to the changes in the joint coordinates); and (ii) the skew-symmetrical part that takes into account changes in the manipulator Jacobian (due to the equilibrium shift caused by the externally applied force) [12, 33, 34]. However, for the preliminary design purposes, the primary interest focuses on the symmetrical part that is evaluated below assuming that effect of the external forces is negligible.

Since matrix $\mathbf{K}_\theta$ is non-singular (it describes the stiffness of the virtual springs), the variable $\delta \boldsymbol{\theta}_i$ can be expressed via $\mathbf{f}_i$ using equations $\mathbf{J}_\theta^{iT} \cdot \mathbf{f}_i = \boldsymbol{\tau}_\theta^i$ and $\boldsymbol{\tau}_\theta^i = \mathbf{K}_\theta \cdot \delta \boldsymbol{\theta}_i$. This yields substitution $\delta \boldsymbol{\theta}_i = (\mathbf{K}_\theta^{-1} \mathbf{J}_\theta^{iT}) \cdot \mathbf{f}_i$ allowing reducing the kinetostatic model to a system of two matrix equations

$$(\mathbf{J}_\theta^i \mathbf{K}_\theta^{-1} \mathbf{J}_\theta^{iT}) \cdot \mathbf{f}_i + \mathbf{J}_q^i \cdot \delta \mathbf{q}_i = \delta \mathbf{t}_i; \qquad \mathbf{J}_q^{iT} \cdot \mathbf{f}_i = \mathbf{0} \tag{10}$$

with unknowns $\mathbf{f}_i$ and $\Delta \mathbf{q}_i$. This system can be also rewritten in a matrix form

$$\begin{bmatrix} \mathbf{S}_\theta^i & \mathbf{J}_q^i \\ \mathbf{J}_q^{iT} & \mathbf{0} \end{bmatrix} \cdot \begin{bmatrix} \mathbf{f}_i \\ \delta \mathbf{q}_i \end{bmatrix} = \begin{bmatrix} \delta \mathbf{t}_i \\ \mathbf{0} \end{bmatrix} \tag{11}$$

where the sub-matrix $\mathbf{S}_\theta^i = \mathbf{J}_\theta^i \mathbf{K}_\theta^{-1} \mathbf{J}_\theta^{iT}$ describes the spring compliance relative to the end-effector, and the sub-matrix $\mathbf{J}_q^i$ takes into account the passive joint influence on the end-effector motions. Therefore, for a separate kinematic chain, the desired stiffness matrix $\mathbf{K}_i$ defining the motion-to-force mapping

$$\mathbf{f}_i = \mathbf{K}_i \cdot \delta \mathbf{t}_i, \qquad (12)$$

can be computed by direct inversion of relevant 10×10 matrix in the left-hand side of (11) and extracting from it the 6×6 sub-matrix with indices corresponding to $\mathbf{S}_\theta^i$. It is also worth mentioning that computing $\mathbf{S}_\theta^i$ requires 6×6 inversions only, since $\mathbf{K}_\theta^{-1} = \mathrm{diag}(K_{ctr}^{-1}, \mathbf{K}_{act}^{-1}, \mathbf{K}_{Foot}^{-1}, \mathbf{K}_{Leg}^{-1})$ and

$$\mathbf{S}_\theta^i = \mathbf{J}_{\theta ctr}^i \cdot K_{ctr}^{-1} \cdot \mathbf{J}_{\theta ctr}^{iT} + \mathbf{J}_{\theta act}^i \cdot \mathbf{K}_{act}^{-1} \cdot \mathbf{J}_{\theta act}^{iT} + \mathbf{J}_{\theta Foot}^i \cdot \mathbf{K}_{Foot}^{-1} \cdot \mathbf{J}_{\theta Foot}^{iT} + \mathbf{J}_{\theta Leg}^i \cdot \mathbf{K}_{Leg}^{-1} \cdot \mathbf{J}_{\theta Leg}^{iT}, \qquad (13)$$

where $\mathbf{J}_{\theta ctr}^i, ... \mathbf{J}_{\theta Leg}^i$ are the corresponding submatrices of the Jacobian $\mathbf{J}_\theta^i$.

Solvability of system (11) in general case, i.e. for any given $\mathbf{J}_\theta^i$ and $\mathbf{J}_q^i$, cannot be proved. Moreover, if the matrix $\mathbf{J}_q^i$ is singular, the passive joint coordinates $\mathbf{q}_i$ cannot be found uniquely. From a physical point of view, it means that if the kinematic chain is located in a singular posture, then certain displacements $\delta \mathbf{t}_i$ can be generated by infinite combinations of the passive joints. But for the variable $\mathbf{f}_i$ the corresponding solution is unique (since the matrix $\mathbf{J}_\theta^i$ is obviously non-singular if at least one 6 dof spring is included in a serial kinematic chain). On the other hand, the singularity may produce an infinite number of stiffness matrices for the same spatial location of the end-effector and for different values $\mathbf{q}_i$ provided by the inverse kinematics. A special technique to tackle this case, based on the singular value decomposition, is presented in appendix A.

After the stiffness matrices $\mathbf{K}_i$ for all kinematic chains are computed, the stiffness of the entire manipulator can be found by simple addition

$$\mathbf{K}_m = \sum_{i=1}^n \mathbf{K}_i \qquad (14)$$

This follows from the superposition principle, because the total external force corresponding to the end-effector displacement $\delta \mathbf{t}$ (the same for all kinematic chains) can be expressed as $\mathbf{f} = \sum_{i=1}^n \mathbf{f}_i$ where $\mathbf{f}_i = \mathbf{K}_i \cdot \delta \mathbf{t}$.

It should be stressed that, for a separate kinematic chain, the stiffness matrix $\mathbf{K}_i$ is not invertible, since some motions of the end-effector do not produce the virtual spring reactions (because of passive joints influence). However, for the entire manipulator, the stiffness matrix $\mathbf{K}_m$ is usually positive definite and invertible for all non-singular postures (with respect to $\mathbf{q}_i$). For example, for the 3-dof translational manipulators presented in Fig.2, $rank(\mathbf{K}_i) = 2$ but $rank(\sum_{i=1}^3 \mathbf{K}_i) = 6$, which ensures the manipulator structure resistance to all possible end-effector displacements.

## 2.5. Comparison with other results

The main advantage of the proposed methodology is its applicability to overconstrained mechanisms. To describe it in details, let us briefly review an alternative technique [41] that was applied to the 3-dof translational manipulator. This technique originates from the same principal equations but the solution strategy of the known method begins from straightforward elimination of the passive joint variables $\mathbf{q}_i$ using the differential kinematic equations (2) only. Obviously, the feasibility of this step depends on the solvability of the equivalent matrix system

$$\begin{bmatrix} \mathbf{I} & -\mathbf{J}_q^1 & & \\ \mathbf{I} & & -\mathbf{J}_q^2 & \\ \mathbf{I} & & & -\mathbf{J}_q^3 \end{bmatrix} \cdot \begin{bmatrix} \delta \mathbf{t} \\ \delta \mathbf{q}_1 \\ \delta \mathbf{q}_2 \\ \delta \mathbf{q}_3 \end{bmatrix} = \begin{bmatrix} \mathbf{J}_\theta^1 & & \\ & \mathbf{J}_\theta^2 & \\ & & \mathbf{J}_\theta^3 \end{bmatrix} \cdot \begin{bmatrix} \delta \boldsymbol{\theta}_1 \\ \delta \boldsymbol{\theta}_2 \\ \delta \boldsymbol{\theta}_3 \end{bmatrix} \qquad (15)$$

where $\delta \mathbf{t}$ and $\delta \mathbf{q}_i$ are treated as unknowns. In the non-constrained case (for the 3-PUU architecture, for instance) the matrix in the left-hand side of (15) is square, of size 18×18. So, it can be inverted usually. However, for overconstrained manipulators, this matrix is non-square, and system (15) cannot be solved uniquely. For example, for manipulators with parallelogram-type legs (Orthoglide, Delta, etc.) the matrix size is 18×15. So, in [31] three additional (virtual) passive joints were introduced to solve the problem. But, obviously, such a modification changes the manipulator architecture and also its stiffness matrix, doubting validity of the corresponding model.

Besides, the proposed method allows computing the stiffness matrix even for the singular manipulator postures and does not incorporate the least-square pseudo-inversions applied by other authors. This is achieved by using another solution strategy, which is applied to for each kinematic chain separately and considers the kinematic and static-equilibrium equations simultaneously. Formal theoretical proof of this feature is based on the singular-value decomposition and is presented in Appendix A, where system (11) is solved for the general case, independent of a relevant manipulator posture (singular or non-singular). In particular, for each kinematic chain, the derived analytical solution allows detecting the subspace of the dimension that defines the motions, which do not cause force/torque reactions of the virtual springs. The advantages of the developed method are presented in Sub-section 4.5, where the stiffness matrix is computed for the "flat" and "bar" singularities of the Orthoglide manipulator. These advantages are also confirmed by the numerical analysis the Orthoglide parallel manipulator presented below.

Some additional conveniences are included in the modeling stage. In particular, the kinematic models of the chains may include several redundant springs that are totally compensated by relevant passive joints. However, there is no need to eliminate these springs from the model manually, since they do not increase the matrix sizes in system (11). This allows including in the model 6-dof virtual springs of general type derived directly from FEA-modeling, without any modifications.

Another advantage of the proposed technique is that it can be generalized easily. Within this paper, it is applied to the stiffness modelling of *n*-dof manipulators with actuators located between the base and the foot. However, it can be easily modified to cover other actuator locations, which may be included in the foot or in the leg. A further generalization is related to the similarity of the kinematic chains. This assumption can be easily relaxed here as it influences on the Jacobian computing only. After the Jacobians are determined, the stiffness matrices for all chains may be computed in the same manner and then aggregated.

## 3. Evaluating model parameters

The adopted stiffness model of each kinematic chain includes four compliant components, which are described by one 1-dof spring corresponding to the actuator control loop and three 6-dof springs corresponding to the actuator transmission and the manipulator links (see Fig. 3). Let us present particular techniques for their evaluation.

### 3.1. *Actuator compliance*

The actuator compliance, described by the scalar parameter $K_{ctr}^{-1}$ and 6×6 matrix $\mathbf{K}_{act}^{-1}$, depends on both the servomechanism mechanics and the control algorithm. Since most modern actuators implement a digital PID control, the main contribution to the compliance is done by the mechanical transmissions. The latter are usually located outside the feedback-control loop and consist of screws, gears, shafts, belts, etc., whose flexibility is comparable with the flexibility of the manipulator links. Because of the complicated mechanical structure of the servomechanisms, these parameters are usually evaluated from static load experiments, by applying the linear regression to the experimental data.

### 3.2. *Link Compliance*

Following a general methodology, the compliance of the manipulator links is described by 6×6 symmetrical positive definite matrices $\mathbf{K}_{Leg}^{-1}$, $\mathbf{K}_{Foot}^{-1}$ corresponding to 6-dof springs with relevant coupling between translational and rotational deformations. This distinguishes our approach from other lumped modeling techniques, where the coupling is neglected and only a subset of deformations is taken into account (presented by several 1-dof springs).

The simplest way to obtain these matrices is to approximate the link by a beam element for which the non-zero elements of the compliance matrix may be expressed analytically:

$$k_{11}=\frac{L}{EA};\ k_{22}=\frac{L^3}{3EI_z};\ k_{33}=\frac{L^3}{3EI_y};\ k_{44}=\frac{L}{GJ};\ k_{55}=\frac{L}{EI_y};\ k_{66}=\frac{L}{EI_z};\ k_{35}=-\frac{L^2}{2EI_y};\ k_{26}=\frac{L^2}{2EI_z} \quad (16)$$

Here $L$ is the link length, $A$ is its cross-section area, $I_y$, $I_z$, and $J$ are the quadratic and polar moments of inertia of the cross-section, and $E$ and $G$ are the Young's and Coulomb's modules respectively.

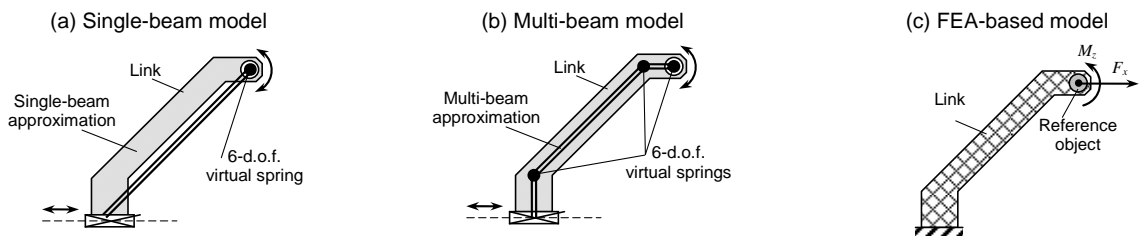

Fig. 4. Evaluation of the stiffness matrix for the Orthoglide foot

However, for certain link shape, the accuracy of the single-beam approximation can be insufficient. In this case, the link can be approximated by a serial chain of the beams, whose compliance is evaluated by applying the same method (i.e. considering the kinematic chain with 6-dof virtual springs, but without passive joints). This leads to the resulting compliance matrix $\mathbf{K}_{Link}^{-1} = \mathbf{J}_b \mathbf{K}_b^{-1} \mathbf{J}_b^T$, where $\mathbf{J}_b$ and $\mathbf{K}_b^{-1}$ incorporate the Jacobian and the compliance matrices for all virtual springs. Examples of the single- and multi-beam approximation of the manipulator foot (for Orthoglide robot) are shown in Fig. 4; corresponding numerical values are presented in Table 1 where they are compared with the FEA modeling results.

TABLE 1.
Comparison of the link stiffness models for the Orthoglide foot

| Method | Compliance Matrix Elements | | | | | |
|---|---|---|---|---|---|---|
|  | $k_{11}$ | $k_{22}$ | $k_{33}$ | $k_{44}$ | $k_{55}$ | $k_{66}$ |
|  | mm/N ×$10^{-4}$ | mm/N ×$10^{-4}$ | mm/N ×$10^{-4}$ | rad/N mm ×$10^{-7}$ | rad/N mm ×$10^{-7}$ | rad/N mm $10^{-7}$ |
| (a) Single-beam approximation | 3.45 | 18.1 | 3.45 | 2.10 | 0.91 | 2.10 |
| (b) Four-beam approximation | 2.77 | 17.9 | 4.34 | 2.11 | 0.91 | 1.95 |
| (c) FEA-based evaluation | 2.45 | 15.9 | 3.24 | 2.07 | 1.71 | 2.06 |

### 3.3. *FEA-based evaluation of stiffness parameters*

For complex link geometries, when the multy-beam approximation is too rough, the most reliable results can be obtained from the FEA modeling. To apply this approach, the CAD model of each link should be extended by introducing an auxiliary 3D object (Fig. 4c), a "pseudo-rigid" body, which is used as a reference for the compliance evaluation. Besides, the link origin must be fixed relative to the global coordinate system. Then, sequentially and separately applying the forces $F_x, F_y, F_z$ and torques $M_x, M_y, M_z$ to the reference object, it is possible to evaluate corresponding linear and angular displacements, which allow computing the stiffness matrix columns.

The main difficulty here is to obtain accurate displacement values by using proper FEA-discretization ("mesh size"). Besides, to increase accuracy, the translational and rotational displacements must be evaluated using the redundant data set describing the reference body motion. For this reason, it is worth applying a dedicated SVD-based algorithm (Appendix B), which allows minimizing the sum of the residual squares. As follows from our study for the Orthoglide robot (Table 1), the single-beam approximation of the Orthoglide foot gives accuracy of about 50%, and the four-beam approximation improves it up to 30% only.

It is worth mentioning that the high computational expenses of FEA is not a critical issue here, because the proposed technique involves only a single evaluation of the link stiffness (in contrast to the straightforward FEA-modeling for the entire manipulator, which requires complete re-computing for each manipulator posture).

## 4. Application examples

To demonstrate the efficiency of the proposed methodology, let us apply it to the comparative stiffness analysis of 3-dof translational mechanisms that employ the Orthoglide architecture [37, 38]. This problem was previously studied using other techniques [31, 40], but the results were essentially different from those obtained from both the FEA-modeling and from the physical experiments. Thus, this section presents several stiffness models for the Orthoglide (Fig. 5) and compares their accuracy with the FEA model of the entire manipulator. It is assumed that influence of the gravity is negligible, since relevant FEA showed very small deflection of the end-platform caused by the weight of the links (less than 0.005 mm).

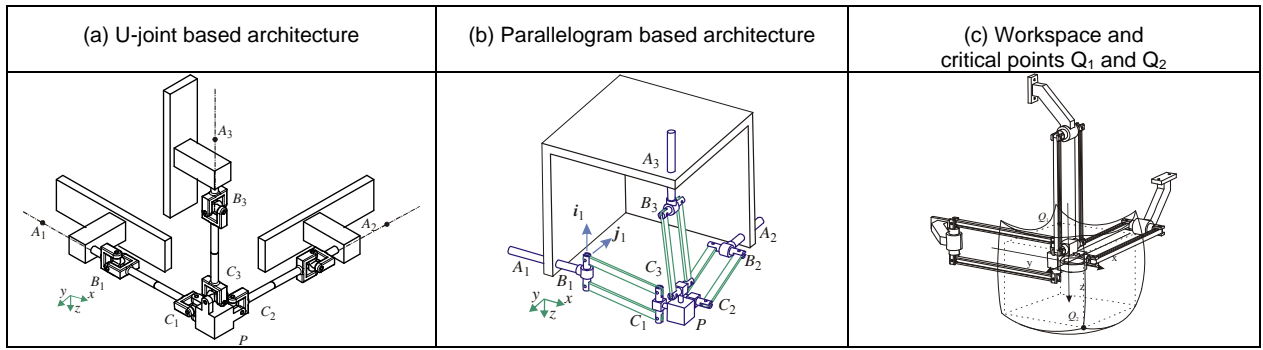

Fig. 5. Kinematics of two 3-dof translational mechanisms employing the Orthoglide architecture

### 4.1. *Manipulator geometry*

The Orthoglide is a Delta-type parallel manipulator dedicated to 3-axis rapid machining applications that was developed to meet the advantages of both serial and parallel kinematic architectures (regular homogeneous workspace with good dynamic performances and stiffness). This manipulator consists of three parallel PRPaR identical chains actuated by three mutually orthogonal linear drives, which are arranged to ensure almost isotropic workspace kinematic properties and to restrict the end-effector motions in translation, with the velocity transmission factors close to 1.0 similar to the conventional XYZ-machines. Moreover, to increase the manipulator stiffness, the kinematic chains impose redundant constraints on the mobile platform (because only two parallelograms would be sufficient to restrict the motion in translation). Hence, this is an over-constrained structure that cannot be evaluated using standard lump modeling techniques.

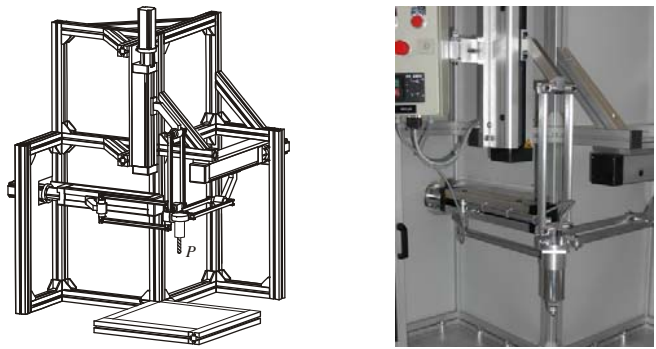

Fig. 6. CAD model of and Orthoglide and its prototype

This architecture was implemented in the Orthoglide prototype (Fig. 6), which was built in Institut de Recherche en Communications et Cybernetique de Nantes (IRCCyN) and satisfies the following design objectives: cubic Cartesian workspace of size 200×200×200 mm, Cartesian velocity and acceleration in the isotropic point 1.2 m/s and 14 m/s2; payload 4 kg; transmission factor range 0.5–2.0. The manipulator kinematics, including the direct and inverse transformations, is described in details in our previous paper [42]. Here we propose the manipulator stiffness model that, in contrast to previous works, does not ignore the over-constraining feature. Also, we compare two alternative architectures, which are kinematically equivalent but differ in the stiffness capabilities.

### 4.2. Stiffness of U-Joint Based Manipulator

First, let us derive the stiffness model for the simplified Orthoglide mechanics, where the legs are comprised of equivalent limbs with U-joints at the ends (Fig 5a). Accordingly, to retain major compliance properties, the limb geometry corresponds to the parallelogram bars with doubled cross-section area.

Let us assume that the world coordinate system is located at the end-effector reference point corresponding to the isotropic manipulator posture (when the legs are mutually perpendicular and parallel to relevant actuator axes). For this assumption, the geometrical models of separate kinematic chains can be described by the expression (1), where $i \in \{x, y, z\}$ and the product components are defined via the standard translational/rotational operators $\mathbf{T}_x(.), \mathbf{T}_y(.), \ldots \mathbf{R}_z(.)$ as follows:

$$\mathbf{T}_{base}^x = \begin{bmatrix} 1 & 0 & 0 & -L-r \\ 0 & 1 & 0 & 0 \\ 0 & 0 & 1 & 0 \\ \hline 0 & 0 & 0 & 1 \end{bmatrix}; \quad \mathbf{T}_{base}^y = \begin{bmatrix} 0 & 0 & 1 & 0 \\ 1 & 0 & 0 & -L-r \\ 0 & 1 & 0 & 0 \\ \hline 0 & 0 & 0 & 1 \end{bmatrix}; \quad \mathbf{T}_{base}^z = \begin{bmatrix} 0 & 1 & 0 & 0 \\ 0 & 0 & 1 & 0 \\ 1 & 0 & 0 & -L-r \\ \hline 0 & 0 & 0 & 1 \end{bmatrix}; \quad (17)$$

$$\mathbf{T}_{tool}^x = \begin{bmatrix} 1 & 0 & 0 & r \\ 0 & 1 & 0 & 0 \\ 0 & 0 & 1 & 0 \\ \hline 0 & 0 & 0 & 1 \end{bmatrix}; \quad \mathbf{T}_{tool}^y = \begin{bmatrix} 0 & 1 & 0 & r \\ 0 & 0 & 1 & 0 \\ 1 & 0 & 0 & 0 \\ \hline 0 & 0 & 0 & 1 \end{bmatrix}; \quad \mathbf{T}_{base}^z = \begin{bmatrix} 0 & 0 & 1 & r \\ 1 & 0 & 0 & 0 \\ 0 & 1 & 0 & 0 \\ \hline 0 & 0 & 0 & 1 \end{bmatrix}; \quad (18)$$

$$\mathbf{V}_a(q_0 + \theta_0) = \mathbf{T}_x(q_0 + \theta_0); \quad \mathbf{T}_{Foot} = \mathbf{I}; \quad \mathbf{T}_{Leg} = \mathbf{T}_x(L) \quad (19)$$

$$\mathbf{V}_s(\theta_1, \ldots \theta_6) = \mathbf{T}_x(\theta_1) \cdot \mathbf{T}_y(\theta_2) \cdot \mathbf{T}_z(\theta_3) \cdot \mathbf{R}_x(\theta_4) \cdot \mathbf{R}_y(\theta_5) \cdot \mathbf{R}_z(\theta_6) \quad (20)$$

$$\mathbf{V}_{u1}(q_1, q_2) \cdot = \mathbf{R}_z(q_1) \cdot \mathbf{R}_y(q_2); \quad \mathbf{V}_{u2}(q_3, q_4) = \mathbf{R}_y(q_3) \cdot \mathbf{R}_z(q_4); \quad (21)$$

Here $L$, $r$ are the manipulator geometrical parameters (the leg length and end-effector offset respectively), and the remaining variables are the same as in equation (1). Because the end-effector of the rigid manipulator is restricted to the translational motions, the nominal values of the passive joint coordinates are subject to the specific constrains, $q_2 + q_3 = 0$ and $q_1 + q_4 = 0$, which are implicitly incorporated in the direct/inverse kinematics. However, the flexible model allows variations all passive joint coordinates around the nominal values, so the jacobians $\mathbf{J}_q$ must be computed for four variables $q_1,\ldots q_4$.

Using the link stiffness parameters obtained by the FEA-modeling (see Appendix C) and applying the proposed methodology, we computed the compliance matrices for three typical manipulator postures, the principal components of which are presented in Table 2. Below, they are compared with the compliance of the parallelogram-based manipulator. It should be noticed that the equivalent model of the UU-leg includes two parallelogram bars with corresponding stiffness matrix $\mathbf{k}_{Bar}/2$.

### 4.3. *Stiffness of Parallelogram Based Manipulator*

Further, let us considerer the 3-PRPaR architecture where the manipulator legs are composed of the kinematic parallelograms (see Fig. 5b), which corresponds to the final design of the Orthoglide prototype. This obviously imposes some additional kinematic constrains compared to the 3-PUU case and should increase the stiffness, but quantitative comparison requires relevant modeling.

Before evaluating the compliance of the entire manipulator, let us derive the stiffness matrix of the parallelogram. Using the adopted notations, the parallelogram equivalent model may be written as

$$\mathbf{T}_{Plg} = \mathbf{R}_y(q_2)\cdot\mathbf{T}_x(L)\cdot\mathbf{R}_y(-q_2)\cdot\mathbf{V}_s(\theta_7,\theta_8,\ldots) \tag{22}$$

where, compared to the 3-PUU case, the third passive joint is eliminated (it is implicitly assumed that $q_3=-q_2$). On the other hand, the original parallelogram may be split into two serial kinematic chains (the "upper" and "lower" ones) that yields the same multi-serial-chain parallel architecture as considered above. Hence, the parallelogram compliance matrix may be derived using the same stiffness modeling technique proposed in this paper.

Assuming that the main contribution to parallelogram compliance is caused by the bar links of the length *L*, the geometry and flexibility of these kinematic chains can be described by the expressions

$$\mathbf{T}_{up} = \mathbf{T}_z(-d/2)\cdot\mathbf{R}_y(q+\Delta q_1^{up})\cdot\mathbf{T}_x(L)\cdot\mathbf{V}_s(\theta_1^{up},\ldots\theta_6^{up})\cdot\mathbf{R}_y(-q+\Delta q_2^{up})\cdot\mathbf{T}_z(d/2) \tag{23}$$

$$\mathbf{T}_{dn} = \mathbf{T}_z(d/2)\cdot\mathbf{R}_y(q+\Delta q_1^{dn})\cdot\mathbf{T}_x(L)\cdot\mathbf{V}_s(\theta_1^{dn},\ldots\theta_6^{dn})\cdot\mathbf{R}_y(-q+\Delta q_2^{dn})\cdot\mathbf{T}_z(-d/2) \tag{24}$$

where *L, d* are the parallelogram geometrical parameters), $\Delta q_1^i, \Delta q_2^i$, $i\in\{up,dn\}$ are the variations of the passive joint coordinates, $q$ is the parallelogram state coordinate defining its "rigid" posture, $\mathbf{V}_s(.)$ describes displacements in the virtual springs, and the sub/superscripts "*up*" and "*dn*" correspond to the upper and lower chain respectively (Fig. 7).

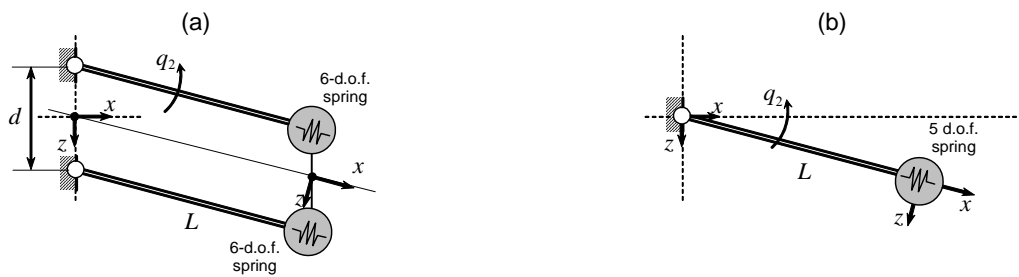

Fig. 7. Lump stiffness model of the parallelogram (a) and its equivalent presentation (b).

Computing Jacobians for the upper chain with respect to $\Delta q_i$ and $\theta_i$ yields

$$\mathbf{J}_q^{up} = \begin{bmatrix} -LS_q + \dfrac{d}{2} & \dfrac{d}{2} \\ 0 & 0 \\ -LC_q & 0 \\ 0 & 0 \\ 1 & 1 \\ 0 & 0 \end{bmatrix}; \quad \mathbf{J}_\theta^{up} = \begin{bmatrix} C_q & 0 & S_q & 0 & \dfrac{d}{2} & 0 \\ 0 & 1 & 0 & -\dfrac{dC_q}{2} & 0 & -\dfrac{dS_q}{2} \\ -S_q & 0 & C_q & 0 & 0 & 0 \\ 0 & 0 & 0 & C_q & 0 & S_q \\ 0 & 0 & 0 & 0 & 1 & 0 \\ 0 & 0 & 0 & -S_q & 0 & C_q \end{bmatrix} \quad (25)$$

where $S_q = \sin(q)$ and $C_q = \sin(q)$. For the lower part, the expressions are similar and differ in signs of $d$ only. Then, defining the bar-link stiffness in the general form as $\mathbf{K}_{Bar} = \begin{bmatrix} K_{ij} \end{bmatrix}$ and performing relevant matrix transformations for both kinematic chains, the parallelogram stiffness matrix is presented analytically as

$$\mathbf{K}_{Plg}(q) = 2 \cdot \begin{bmatrix} K_{11} & 0 & 0 & 0 & 0 & 0 \\ 0 & K_{22} & 0 & 0 & 0 & K_{26} \\ 0 & 0 & 0 & 0 & 0 & 0 \\ \hline 0 & 0 & 0 & K_{44} + \dfrac{d^2 C_q^2 K_{22}}{4} & 0 & \dfrac{d^2 S_{2q} K_{22}}{8} \\ 0 & 0 & 0 & 0 & \dfrac{d^2 C_q^2 K_{11}}{4} & 0 \\ 0 & K_{26} & 0 & \dfrac{d^2 S_{2q} K_{22}}{8} & 0 & K_{66} + \dfrac{d^2 S_q^2 K_{22}}{4} \end{bmatrix} \quad (26)$$

where it is assumed that the x-axis is directed along to the L-links (i.e., similar to the 3-PUU case). It is also worth mentioning that the stiffness parameters $K_{33}$ and $K_{55}$ (which depend on $I_y$, see equations (16)) are completely compensated by the passive joints, and there is no translational stiffness in the z-direction. Moreover, the rotational stiffness around z-axis is defined by the parameter $K_{11}$ (describing the bar compression/tension).

More detailed analysis of the matrix $\mathbf{K}_{Plg}(0)$ at the isotropic point using expressions (16) shows that most of the elements are doubled compared to the stiffness of a single beam. But the rotation about the z-axis (matrix elements $K_{55}$, where the term $L/EI_y$ is replaced by $d^2 EA/2L$) demonstrates essential increase of the stiffness. Numerically, it can be evaluated as the ratio $d^2 A/(8I_y)$, which for the rectangular cross-section of size $b \times h$ is reduced to $1.5\,(d/h)^2$ where the dimension $h$ corresponds to the axis y. Besides, there is some stiffness increase in the x-axis rotation (element $K_{44}$) but it is not so high. For instance, for the considered case study (Orthoglide prototype), the increase in $K_{55}$ is about 25 times, while the element $K_{44}$ increases roughly by 10%. However, for the entire manipulator, three parallelograms yield essential stiffness increase for all rotational axes.

It worth mentioning that, for such description of the parallelogram stiffness, the kinematic chains of 3-PRPaR manipulator are described by the same equations as in the 3-PUU case, but the joint variable $q_3^i$ is not treated as an independent one, since $q_2^i + q_3^i = 0$. The latter must be taken account while computing the jacobians $\mathbf{J}_q^i$ which size is reduced to 6x3 and corresponds to three passive joints $q_1^i, q_2^i, q_4^i$.

To avoid the rank-deficiency of matrix (26) that is produced by the third raw and column, it is necessary to eliminate the third translational spring. This allows the matrix inversion while computing $\mathbf{S}_\theta^i$ (see equations (13)) and reduces the parallelogram stiffness model to a 5-dof virtual spring that includes two translational and three rotational components with relevant coupling between them. Another way to avoid the above singularity is to introduce an arbitrary "fictitious" stiffness at the intersection of the zero raw and column of (26), since its influence will be totally compensated by the corresponding passive joint presented in the Jacobian $\mathbf{J}_q^i$. The latter approach allows minimizing preliminary analytical derivation and replacing them by numerical computations.

Using this model and applying the proposed technique, we computed the compliance matrices for three typical non-singular manipulator postures (Table 2) and also the stiffness matrixes for two singular postures (Table 3). As follows from the comparison with the U-joint case, within the dexterous workspace, the parallelograms allow multiplying the rotational stiffness roughly by 10. This justifies application of this architecture in the Orthoglide prototype design [37].

TABLE 2
Comparison of translational and rotational compliance for 3-PUU and 3-PRPaR manipulators

| MANIPULATOR ARCHITECTURE | Point $Q_0$ $x, y, z = 0.00$ mm | | Point $Q_1$ $x, y, z = -73.65$ mm | | Point $Q_2$ $x, y, z = +126.35$ mm | |
|---|---|---|---|---|---|---|
| | $k_{tran}$ [mm/N] | $k_{rot}$ [rad/N·mm] | $k_{tran}$ [mm/N] | $k_{rot}$ [rad/N·mm] | $k_{tran}$ [mm/N] | $k_{rot}$ [rad/N·mm] |
| 3-PUU manipulator | $2.78 \cdot 10^{-4}$ | $20.9 \cdot 10^{-7}$ | $10.9 \cdot 10^{-4}$ | $24.1 \cdot 10^{-7}$ | $71.3 \cdot 10^{-4}$ | $25.8 \cdot 10^{-7}$ |
| 3-PRPaR manipulator | $2.78 \cdot 10^{-4}$ | $1.94 \cdot 10^{-7}$ | $9.86 \cdot 10^{-4}$ | $2.06 \cdot 10^{-7}$ | $21.2 \cdot 10^{-4}$ | $2.65 \cdot 10^{-7}$ |

### 4.4. Accuracy of the proposed model

To validate the developed stiffness modeling technique, the 3-PRPaR Orthoglide prototype was evaluated using several different lump models and the FEA-based model. The obtained results are presented in Table 4. As follows from them, accuracy of the previous methods [31] is about 25…30%. In contrast, the proposed model that uses 6-dof virtual springs with FEA-based evaluation of the link stiffness parameters, gives accuracy of about 5% for almost all workspace points. The only exceptions are the manipulator configurations that are close to the "flat singularity" (point $Q_2$), when the error for the rotational stiffness rises up to 20%.

This motivated development of the extended stiffness model for the parallelogram legs, which takes into account flexibility of the "axes" corresponding to the links $d$ (see Fig. 7). However, while it yielded essential improvement in accuracy for the translational stiffness, the accuracy of the rotational components in the neighborhood of $Q_2$ was about 15% only. The latter motives further research that takes into account the joint stiffness (in addition to the links and actuators). However, in general, these results confirm advantages of the developed technique and justify its application for the pre-design stage.

TABLE 3
Stiffness matrices $K_{tran}$ of 3-PUU and 3-PRPaR manipulators in singular configurations

| MANIPULATOR ARCHITECTURE | "Flat" singularity $x, y, z = -L/\sqrt{6}$ 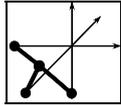 $\mathbf{K}_{tran}$, [N/mm] | "Flat" singularity $x, y, z = +L/\sqrt{3}$ 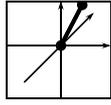 $\mathbf{K}_{tran}$, [N/mm] |
|---|---|---|
| 3-PUU manipulator | $\mathbf{K}_{tran} = \begin{bmatrix} 1.48 & -0.74 & -0.74 \\ -0.74 & 1.48 & -0.74 \\ -0.74 & -0.74 & 1.48 \end{bmatrix} \cdot 10^3$ $rank(\mathbf{K}_{tran}) = 2$ | $\mathbf{K}_{tran} = \begin{bmatrix} 1.78 & 1.78 & 1.78 \\ 1.78 & 1.78 & 1.78 \\ 1.78 & 1.78 & 1.78 \end{bmatrix} \cdot 10^3$ $rank(\mathbf{K}_{tran}) = 1$ |
| 3-PRPaR manipulator | $\mathbf{K}_{tran} = \begin{bmatrix} 1.54 & -0.77 & -0.77 \\ -0.77 & 1.54 & -0.77 \\ -0.77 & -0.77 & 1.54 \end{bmatrix} \cdot 10^3$ $rank(\mathbf{K}_{tran}) = 2$ | $\mathbf{K}_{tran} = \begin{bmatrix} 4.65 & 4.65 & 4.65 \\ 4.65 & 4.65 & 4.65 \\ 4.65 & 4.65 & 4.65 \end{bmatrix} \cdot 10^3$ $rank(\mathbf{K}_{tran}) = 1$ |

TABLE 4
Comparison of compliance modelling results for 3-PRPaR Orthoglide

| STIFFNESS MODEL | Point $Q_0$ $x,y,z = 0.00$ mm | | Point $Q_1$ $x,y,z = -73.65$ mm | | Point $Q_2$ $x,y,z = +126.35$ mm | |
|---|---|---|---|---|---|---|
| | $k_{tran}$ [mm/N] | $k_{rot}$ [rad/N·mm] | $k_{tran}$ [mm/N] | $k_{rot}$ [rad/N·mm] | $k_{tran}$ [mm/N] | $k_{rot}$ [rad/N·mm] |
| Lump model of F.Majou et al. (2007) with additional passive joints | $3.68 \cdot 10^{-4}$ | $2.77 \cdot 10^{-7}$ | $13.8 \cdot 10^{-4}$ | $2.77 \cdot 10^{-7}$ | $34.3 \cdot 10^{-4}$ | $2.78 \cdot 10^{-7}$ |
| Modified model of F.Majou et al. (2007) without additional passive joints | $3.68 \cdot 10^{-4}$ | $1.26 \cdot 10^{-7}$ | $12.5 \cdot 10^{-4}$ | $1.26 \cdot 10^{-7}$ | $24.7 \cdot 10^{-4}$ | $1.26 \cdot 10^{-7}$ |
| Over-constrained lump model with 6-dof springs | $2.78 \cdot 10^{-4}$ | $1.94 \cdot 10^{-7}$ | $9.86 \cdot 10^{-4}$ | $2.06 \cdot 10^{-7}$ | $21.2 \cdot 10^{-4}$ | $2.65 \cdot 10^{-7}$ |
| Extended over-constrained model with 6-dof springs | $2.93 \cdot 10^{-4}$ | $2.02 \cdot 10^{-7}$ | $10.2 \cdot 10^{-4}$ | $2.15 \cdot 10^{-7}$ | $21.9 \cdot 10^{-4}$ | $2.76 \cdot 10^{-7}$ |
| FEA-based model | $3.05 \cdot 10^{-4}$ | $2.05 \cdot 10^{-7}$ | $10.9 \cdot 10^{-4}$ | $2.17 \cdot 10^{-7}$ | $26.8 \cdot 10^{-4}$ | $2.67 \cdot 10^{-7}$ |

## 5. Conclusions

The paper proposes a new systematic method for computing the stiffness matrix of overconstrained parallel manipulators. It is based on a multidimensional lumped model of the flexible links, whose parameters are evaluated via the FEA-modeling and describe both the translational/rotational compliances and the coupling between them. In contrast to previous works, the method employs a new solution strategy of the kinetostatic equations, which considers simultaneously the kinematic and static relations for each separate kinematic chain and then aggregates the partial solutions in a total one. This allows computing the stiffness matrices for overconstrained mechanisms for any given manipulator posture, including singular configurations and their neighborhood. Another advantage is the computational simplicity that requires low-dimensional matrix inversion compared to other techniques. Besides, the method does not require manual elimination of the redundant spring corresponding to the passive joints, since this operation is inherently included in the numerical algorithm. Using the proposed methodology, we also derived the analytical 5-dof stiffness model of the parallelogram-based link.

The efficiency and accuracy of the proposed method was demonstrated through application examples that compare the stiffness of two parallel manipulators of the Orthoglide family (with U-joint based and parallelogram based links). Relevant simulation results have confirmed essential advantages of the parallelogram based architecture and validated adopted design of the Orthoglide prototype. Accuracy of the proposed model was evaluated via comparison with FEA modeling.

While applied to mechanisms with similar kinematic chains and actuators located between the base and foot, the method can be extended to other parallel architectures to cover different actuator locations and dissimilar chain geometry. So, future work will focus on the stiffness modeling of more complicated parallel mechanisms (such as the Verne machine) and also on the experimental verification of the stiffness models for the Orthoglide robot. Another prospective research direction is the stiffness analysis of heavy manipulators, for which the influence of gravity is essential and cannot be neglected.

## Acknowledgments

This work has been partially funded by the European projects NEXT, acronyms for "Next Generation of Productions Systems", Project no° IP 011815.

# Appendices

*Appendix A. SVD-based computing of stiffness matrix for a kinematic chain with passive joints*

Let us consider the kinetostatic model of a separate kinematic chain

$$\mathbf{S}_\theta \cdot \mathbf{f} + \mathbf{J}_q \cdot \delta\mathbf{q} = \delta\mathbf{t}; \quad \mathbf{J}_q^T \cdot \mathbf{f} = \mathbf{0}$$

which defines portion of the external force $\mathbf{f} \in R^6$ and the variations of the passive joint coordinates $\delta\mathbf{q} \in R^m$ corresponding to the end-effector motion $\delta\mathbf{t} \in R^6$ (for convenience, the chain-number indices *i* are omitted). Here *m* is the passive joint number, $\mathbf{J}_q$ is the passive joint jacobian of size 6xm, and $\mathbf{S}_\theta$ is the 6x6 positive-definite symmetric matrix of the virtual springs compliance relative to the end-effector (see Section 2 for details).

To find the desired mapping $\mathbf{f} = \mathbf{K} \cdot \delta\mathbf{t}$, let us apply the SVD decomposition to the jacobian $\mathbf{J}_q$ that yields to the following factorisation

$$\mathbf{J}_q = \mathbf{U}_q \cdot \mathbf{\Sigma}_q \cdot \mathbf{V}_q^T$$

where $\mathbf{U}_q$ and $\mathbf{V}_q$ are the orthogonal matrices of the size 6×6 and m×m respectively (i.e., $\mathbf{U}_q^T \mathbf{U}_q = \mathbf{I}$, $\mathbf{V}_q^T \mathbf{V}_q = \mathbf{I}$), and $\mathbf{\Sigma}_q$ is the 6×m quasi-diagonal matrix with non-negative elements $\sigma_1, \ldots \sigma_m$ (singular values). Then, after substitution $\mathbf{J}_q$ and left-multiplication of the first equation by $\mathbf{U}_q^T$ and the second one by $\mathbf{V}_q^T$, the original system may be rewritten as

$$\mathbf{U}_q^T \mathbf{S}_\theta \mathbf{U}_q \cdot (\mathbf{U}_q^T \mathbf{f}) + \mathbf{\Sigma}_q \cdot (\mathbf{V}_q^T \delta\mathbf{q}) = \mathbf{U}_q^T \delta\mathbf{t}; \quad \mathbf{\Sigma}_q^T \cdot (\mathbf{U}_q^T \mathbf{f}) = \mathbf{0}.$$

The latter may be treated as the orthogonal linear transformations of the variables

$$\delta\mathbf{t}' = \mathbf{U}_q^T \delta\mathbf{t}; \quad \mathbf{f}' = \mathbf{U}_q^T \mathbf{f}; \quad \delta\mathbf{q}' = \mathbf{V}_q^T \delta\mathbf{q}$$

with respect to which the considered system is simplified down to:

$$\mathbf{U}_q^T \mathbf{S}_\theta \mathbf{U}_q \cdot \mathbf{f}' + \mathbf{\Sigma}_q \cdot \delta\mathbf{q}' = \delta\mathbf{t}'; \quad \mathbf{\Sigma}_q^T \cdot \mathbf{f}' = \mathbf{0}.$$

Further, because the matrix $\mathbf{\Sigma}_q$ is quasi-diagonal, the second equation may be re-written in a scalar from as

$$\sigma_k \cdot f_k' = 0, \quad k = 1, \ldots m,$$

which gives the trivial solutions for the first *r* components of the variable $\mathbf{f}'$

$$f_k' = 0, \quad k = 1, \ldots r$$

corresponding to $\sigma_k \neq 0$ where $k = 1, \ldots r$ and $r = rank(\mathbf{J}_q)$ (usually, $r = m$ within the dexterous workspace, and only for some singular configurations $r < m$.). The remaining (*n-r*) components of $\mathbf{f}'$ corresponding to $\sigma_k = 0$ can be found from the portion of the first matrix equation which in the scalar form is presented as

$$\sum_{l=r+1}^{6} \mathbf{u}_q^{kT} \mathbf{S}_\theta \mathbf{u}_q^k \cdot f'_{lk} = \delta t'_k; \quad k = r+1, \ldots 6$$

where $\mathbf{u}_q^k \in R^6$, $k = 1, \ldots 6$ are the orthogonal vector-columns comprising the matrix $\mathbf{U}_q$. Hence, in the matrix form, the expression for the vector variable $\mathbf{f}'$ can be written as

$$\mathbf{f}' = \begin{bmatrix} \mathbf{0}_{r \times r} & \mathbf{0}_{r \times (6-r)} \\ \mathbf{0}_{(6-r) \times r} & (\mathbf{U}_q^{dT} \mathbf{S}_\theta \mathbf{U}_q^d)^{-1} \end{bmatrix} \cdot \delta \mathbf{t}'$$

where $\mathbf{U}_q^d = [\mathbf{u}_q^{r+1}, \ldots \mathbf{u}_q^6]$ is a "rank-deficient" part of the matrix $\mathbf{U}_q$ obtained by its partitioning $\mathbf{U}_q = [\mathbf{U}_q^r, \mathbf{U}_q^d]$ into two submatrices of the size $6 \times r$ and $6 \times (6-r)$ respectively:

Thus, after restoring the original variables $\mathbf{f}$, $\delta \mathbf{q}$ and left-multiplication by $\mathbf{U}_q$, the above relation is transformed into

$$\mathbf{f} = \begin{bmatrix} \mathbf{U}_q^r & \mathbf{U}_q^d \end{bmatrix} \cdot \begin{bmatrix} \mathbf{0} & \mathbf{0} \\ \mathbf{0} & (\mathbf{U}_q^{dT} \mathbf{S}_\theta \mathbf{U}_q^d)^{-1} \end{bmatrix} \cdot \begin{bmatrix} \mathbf{U}_q^{rT} \\ \mathbf{U}_q^{dT} \end{bmatrix} \cdot \delta \mathbf{t}$$

which after simplification yields the final expression for the stiffness of a kinematic chain

$$\mathbf{K} = \mathbf{U}_q^d (\mathbf{U}_q^{dT} \mathbf{S}_\theta \mathbf{U}_q^d)^{-1} \mathbf{U}_q^{dT}$$

where the matrix $\mathbf{K}$ is obviously conservative.

In this expression, the matrix $\mathbf{S}_\theta = \mathbf{J}_\theta \mathbf{K}_\theta^{-1} \mathbf{J}_\theta^T$ describes the spatial location and compliance of the virtual springs, and the matrix $\mathbf{U}_q^d$ characterizes impact of the passive joints, which slacken the springs effect by accepting certain motions without the force/torque reactions. The rank of the obtained matrix $rank(\mathbf{K}) = 6 - r$ depends on the number of passive joints $m$ and the kinematic chain posture ($r \leq m$). Apparently, for the case without passive joints, the expression for $\mathbf{k} = \mathbf{K}^{-1}$ is reduced to the known one, i.e. $\mathbf{k} = \mathbf{J}_\theta \mathbf{K}_\theta^{-1} \mathbf{J}_\theta^T$.

**Appendix B**. *CAD-based computing of the link compliance matrix*

Let us assume that the FEA-modelling provided six data sets describing the displacement of the reference object caused by successive applications of the forces $F_x, F_y, F_z$ and the torques $M_x, M_y, M_z$ along the axes of the virtual spring coordinate system. Each such data set may be formally described as $\{\mathbf{p}_k, \mathbf{d}_k \mid k=1, 2, \ldots m\}$ where $\mathbf{p}_k$ and $\mathbf{d}_k$ are respectively the Cartesian position and the Cartesian displacement of the *k*th node in the link-base coordinate system.

To evaluate the reference object translation $(\Delta p_x, \Delta p_y, \Delta p_z)$ and rotation $(\Delta \varphi_x, \Delta \varphi_y, \Delta \varphi_z)$ relative to the virtual spring centre $\mathbf{p}_0$, let us fit the data by the model

$$\mathbf{d}_k = \mathbf{R}(\mathbf{p}_k - \mathbf{p}_0) + \mathbf{t} - (\mathbf{p}_k - \mathbf{p}_0),$$

which includes, as the parameters, the translation vector $\mathbf{t}$ and the orthogonal rotation matrix $\mathbf{R}$ of sizes 3x1 and 3x3 respectively. After defining $\mathbf{g}_k = \mathbf{p}_k - \mathbf{p}_0$ and $\mathbf{g}'_k = \mathbf{p}_k - \mathbf{p}_0 + \mathbf{d}_k$, the model may be rewritten in the form

$$\mathbf{g}'_k = \mathbf{R} \cdot \mathbf{g}_k + \mathbf{t}; \quad k = 1, \ldots m,$$

which is known in the matrix analysis as the "Procrustes problem" and admits the minimum least square solution [43]

$$R = \mathbf{V}\mathbf{U}^T; \quad \mathbf{t} = m^{-1}\sum_{k=1}^{m}\mathbf{g}'_k - m^{-1}\mathbf{R}\sum_{k=1}^{m}\mathbf{g}_k$$

via the SVD factorization of the following 3x3 matrix

$$\sum_{k=1}^{m}\mathbf{g}_k\mathbf{g}'^T_k = \mathbf{U}_g \cdot \mathbf{\Sigma}_g \cdot \mathbf{V}_g^T$$

where $\mathbf{U}_g$ and $\mathbf{V}_g$ are the orthogonal matrices of the size 3×3, and $\mathbf{\Sigma}_g$ is the 3x3 diagonal matrix (positive definite, if m ≥ 3). For small displacements $\mathbf{d}_k$, the rotation matrix may be rewritten in the differential from, which gives explicit expressions for the rotation angles:

$$\Delta\varphi_x = r_{23}; \quad \Delta\varphi_y = r_{13}; \quad \Delta\varphi_z = r_{12}.$$

The translational displacements are extracted from the vector $\mathbf{t}$:

$$\Delta p_x = t_1; \quad \Delta p_y = t_2; \quad \Delta p_z = t_3.$$

Then the obtained values $\Delta p_x, \ldots, \Delta\varphi_z$ are scaled by dividing by the corresponding force/torque amplitude. And, after applying this algorithm to all six data sets (corresponding to $F_x, F_y, \ldots M_z$), the desired compliance matrix of size 6x6 is constructed as

$$\mathbf{k}_{CAD} = \begin{bmatrix} \Delta p_x(F_x)/F_x & \Delta p_x(F_y)/F_y & \cdots & \Delta p_x(M_z)/M_z \\ \Delta p_y(F_x)/F_x & \Delta p_y(F_y)/F_y & \cdots & \Delta p_y(M_z)/M_z \\ \vdots & \vdots & \cdots & \vdots \\ \Delta\varphi_z(F_x)/F_x & \Delta\varphi_z(F_y)/F_y & \cdots & \Delta\varphi_z(M_z)/M_z \end{bmatrix}$$

where $\Delta p_x(F_y)$ denotes the displacement $\Delta p_x$ caused by the applied force $\Delta F_y$, etc. Finally, to compensate some small computational errors, the obtained matrix is symmetrized

$$\mathbf{k}_{link} = (\mathbf{k}_{CAD} + \mathbf{k}^T_{CAD})/2$$

by averaging the non-diagonal symmetrical elements.

**Appendix C**. *Compliance parameters of the Orthoglide links*

For the Orthoglide manipulator, the actuator compliance was evaluated as $k_{ctr} = 10^{-5} mm/N$ while the links compliance matrices were computed via the FEA-based simulation using technique presented in Appendix 2, which yielded:

$$\mathbf{k}_{Foot} = \left[\begin{array}{cccccc} 2.45 \cdot 10^{-4} & -2.73 \cdot 10^{-4} & 0 & 0 & 0 & -5.48 \cdot 10^{-6} \\ -2.73 \cdot 10^{-4} & 3.24 \cdot 10^{-4} & 0 & 0 & 0 & 7.04 \cdot 10^{-6} \\ 0 & 0 & 1.59 \cdot 10^{-3} & 9.90 \cdot 10^{-6} & -1.27 \cdot 10^{-5} & 0 \\ \hline 0 & 0 & 9.90 \cdot 10^{-6} & 2.07 \cdot 10^{-7} & 0 & 0 \\ 0 & 0 & -1.27 \cdot 10^{-5} & 0 & 2.06 \cdot 10^{-7} & 0 \\ -5.48 \cdot 10^{-6} & 7.04 \cdot 10^{-6} & 0 & 0 & 0 & 1.71 \cdot 10^{-7} \end{array}\right]$$

$$\mathbf{k}_{Bar} = \left[\begin{array}{cccccc} 4.50 \cdot 10^{-5} & 0 & 0 & 0 & 0 & 0 \\ 0 & 8.01 \cdot 10^{-2} & 0 & 0 & 0 & 3.98 \cdot 10^{-4} \\ 0 & 0 & 3.64 \cdot 10^{-2} & 0 & -1.71 \cdot 10^{-4} & 0 \\ \hline 0 & 0 & 0 & 3.76 \cdot 10^{-6} & 0 & 0 \\ 0 & 0 & -1.71 \cdot 10^{-4} & 0 & 1.09 \cdot 10^{-6} & 0 \\ 0 & 3.98 \cdot 10^{-4} & 0 & 0 & 0 & 2.65 \cdot 10^{-6} \end{array}\right]$$

$$\mathbf{k}_{Axis} = \left[\begin{array}{cccccc} 1.99 \cdot 10^{-6} & 0 & 0 & 0 & 0 & 0 \\ 0 & 1.29 \cdot 10^{-5} & 0 & 0 & 0 & 2.61 \cdot 10^{-7} \\ 0 & 0 & 1.50 \cdot 10^{-5} & 0 & -7.64 \cdot 10^{-7} & 0 \\ \hline 0 & 0 & 0 & 6.81 \cdot 10^{-8} & 0 & 0 \\ 0 & 0 & -7.64 \cdot 10^{-7} & 0 & 8.23 \cdot 10^{-8} & 0 \\ 0 & 2.61 \cdot 10^{-7} & 0 & 0 & 0 & 2.67 \cdot 10^{-8} \end{array}\right];$$

$$\mathbf{k}_{Act} = \left[\begin{array}{cccccc} 1.88 \cdot 10^{-6} & 0 & 0 & 0 & 0 & 0 \\ 0 & 3.83 \cdot 10^{-7} & 0 & 0 & 0 & 0 \\ 0 & 0 & 9.99 \cdot 10^{-6} & 2.90 \cdot 10^{-7} & -0.45 \cdot 10^{-7} & 0 \\ \hline 0 & 0 & 2.90 \cdot 10^{-7} & 1.55 \cdot 10^{-8} & 0 & 0 \\ 0 & 0 & -0.45 \cdot 10^{-7} & 0 & 5.19 \cdot 10^{-10} & 0 \\ 0 & 0 & 0 & 0 & 0 & 4.86 \cdot 10^{-10} \end{array}\right]$$

In these matrices, the units are [mm], [rad], [N] and [N×mm] for the length, angle, force and torque respectively. As following from these results, the most compliant manipulator components are the foots and the parallelogram bars, while the remaining elements have rigidity of 5 – 10 times higher.